\documentclass[letterpaper, 10 pt, journal, twoside]{ieeetran}
                                   
\usepackage{amsfonts}
\usepackage{mathrsfs}
\usepackage{colortbl}
\usepackage{amssymb,amsmath,graphicx}
\usepackage{subfigure}
\usepackage{cite}
\usepackage{flushend}

\newtheorem{Problem}{\bf Problem}
\newtheorem{Definition}{\bf Definition}

\newtheorem{Theorem}{\bf Theorem}

\newtheorem{Remark}{\bf Remark}
\newtheorem{Property}{\bf Property}
\newtheorem{Proposition}{\bf Proposition}

\renewcommand{\c}{\mathbf{c}}
\renewcommand{\u}{\mathbf{u}}
\newcommand{\x}{\mathbf{x}}
\newcommand{\R}{\mathbb{R}}
\newcommand{\0}{\mathbf{0}}
\newcommand{\s}{\mathrm{s}}
\newcommand{\X}{\mathcal{X}}
\newcommand{\U}{\mathcal{U}}
\newcommand{\p}{\mathcal{P}}
\newcommand{\ddt}{\frac{ \rm d}{ {\rm d} t}}
\newcommand{\xxi}{\boldsymbol{\xi}}
\newcommand{\eeta}{\boldsymbol{\eta}}

\title{
Instantaneous Capture Input for Balancing the Variable Height Inverted Pendulum
}

\author{Junwei Liu$^{1}$, Hua Chen$^{2}$, Patrick M. Wensing$^{3}$, and Wei Zhang$^{2}$
\thanks{Manuscript received: February 24, 2021; Revised May 27, 2021; 
Accepted June 27, 2021.}
\thanks{This paper was recommended for publication by Editor Abderrahmane Kheddar upon evaluation of the Associate Editor and Reviewers' comments. This work was supported in part by National Natural Science Foundation of China under Grant No. 62073159 and Grant No. 62003155, in part by the Shenzhen Science and Technology Program under Grant No. JCYJ20200109141601708, and in part by the Science, Technology and Innovation Commission of Shenzhen Municipality under grant no. ZDSYS20200811143601004. \emph{(Corresponding author: Wei Zhang.)}}
\thanks{$^{1}$Junwei Liu is with Department of Mechanical and Energy Engineering, Southern University of Science and Technology, Shenzhen 518055, China 
{\tt\small liujw@sustech.edu.cn}}
\thanks{$^{2}$Hua Chen and Wei Zhang are with Shenzhen Key Laboratory of Biomimetic Robotics and Intelligent Systems, Department of Mechanical and Energy Engineering, Southern University of Science and Technology, Shenzhen, 518055, China 
{\tt\small \{chenh6,zhangw3\}@sustech.edu.cn}}
\thanks{$^{3}$P. M. Wensing is with the Department of Aerospace and Mechanical Engineering, University of Notre Dame, Notre Dame, IN 46556, USA 
{\tt\small pwensing@nd.edu}}
}

\begin{document}
\maketitle

\begin{abstract}
Balancing is a fundamental need for legged robots due to their unstable floating-base nature. Balance control has been thoroughly studied for simple models such as the linear inverted pendulum thanks to the concept of the instantaneous capture point (ICP), yet the constant center of mass height assumption limits the application. This paper explores balancing of the variable-height inverted pendulum (VHIP) model by introducing the \emph{instantaneous capture input} (ICI), an extension of the ICP based on its key properties. Namely, the ICI can be computed as a function of the state, and when this function is used as the control policy, the ICI is rendered stationary and the system will eventually come to a stop. This characterization induces an analytical region of capturable states for the VHIP, which can be used to conceptually guide where to step. To further address state and control constraints during recovery, we present and theoretically analyze an explicit ICI-based controller with online optimal feedback gains. Simulations demonstrate the validity of our controller for capturability maintenance compared to an approach based on the divergent component of motion.
\end{abstract}


\section{INTRODUCTION}

\IEEEPARstart{O}{ne} of the main reasons preventing the wider use of humanoid robots is the challenge of meeting  safety requirements, such as balancing without falling.
During the past few decades, various stability concepts have been extensively used in legged locomotion, including conditions based on the Poincar\'e map \cite{Grizzle14}, viability theory
\cite{Wieber02,Zaytsev18}, capturability analysis
\cite{Koolen12}, and so on. Among them, capturability measures the ability of a legged robot to come to a stable stop, which has recently received increasing attention because of its numerical tractability for analysis \cite{Posa17} and control implementation \cite{Pratt12}. In real-world scenarios, rather than simply coming to a stop, achieving desired body motions or setpoints is critical to perform assigned tasks. This more general problem is commonly referred to as the problem of balance control \cite{Koolen16,Morisawa12}. To gain insight into implementation, a widely accepted pipeline for the analysis of balancing is to adopt simplified models, and in particular, models within the class of inverted pendulums.

Considerable advances have been achieved in balancing using the linear inverted pendulum (LIP) model \cite{Kajita91,Kajita01}, with the instantaneous capture point (ICP) concept \cite{Pratt06} playing a key role. The ICP is a function of the LIP state that defines a point on the ground. If the center of pressure (CoP) were placed at the ICP then, 1) the ICP would remain constant, and 2) the LIP would eventually come to a stop.
The ICP concept was independently discovered in other literature \cite{Hof05,Takenaka09} with its other properties illuminated therein. Further, capturability of the LIP is fully determined by the ICP \cite{Koolen12}: the ICP must lie within the support polygon for the LIP to be capturable without changing its support. The balance control problem, on the other hand, can be tackled with ICP feedback controllers \cite{Englsberger15,Morisawa12}, while paying special attention to keep the ICP inside the support polygon for maintaining capturability.
The ICP notion has also been extended to derive capturability constraints for walking pattern generation \cite{Pajon19} as well as multicontact balancing \cite{Prete18}.

To relax the linear CoM motion assumption, research efforts have been directed toward a variable-height inverted pendulum (VHIP) model. Whereas the LIP model implicitly assumes a fixed virtual leg stiffness, the VHIP model allows the virtual leg stiffness to change. 
In \cite{Koolen16}, a characterization of capturable states was identified for the case where the support polygon shrinks to a point and the virtual leg stiffness is allowed to take any nonnegative value.
Posa et al. \cite{Posa17} applied a sums-of-squares optimization method to approximately compute capturable states of the VHIP. Other work has extended the capture point technique to accommodate height variations in \cite{Hofslot19} and \cite{Ramos15}. 

A separate group of work has considered extensions of the ICP dynamics to a 3D setting through the concept of the divergent component of motion (DCM) \cite{Hopkins14,Caron18,Caron20a,Caron20b}. 
In particular, a boundedness condition was derived to characterize feasible inputs that ensure capturability \cite{Caron20a}, and a DCM feedback controller was employed to deal with the balance control problem \cite{Caron20b}.
Despite the progress made in balancing with the VHIP model, several problems remain open compared to the LIP. The VHIP still lacks an analytical characterization of its capturable states, especially for those with nonzero vertical velocities. Moreover, the treatment of constraints in any such analysis remains a hurdle due to the nonlinear dynamics of the VHIP, which hinders its transition to practice.

To fill these gaps, we introduce a new concept called the instantaneous capture input (ICI). We use this concept to characterize capturability and to synthesize feedback controllers for balance control of the VHIP. Our main contributions are threefold: 
First, the new ICI concept extends the ICP concept by considering capturability through the lens of the VHIP 
control input. Indeed, the ICP can be applied as a fixed control input for the CoP of the LIP to ensure balance, and a generalization of this property has not yet appeared for the VHIP. The benefits of this concept over previous DCM extensions are discussed.
Second, the proposed ICI enables a tight characterization of the capturable states of the VHIP with a closed-form expression. Moreover, the region uncovers that a fixed contact point and virtual leg stiffness conceptually suffices for the VHIP to balance with height variation, largely enriching the motion for balancing based on the ICP \cite{Pratt06,Koolen12}.
Third, the proposed controller extends the ICP-based controller presented in \cite{Englsberger15} by incorporating nonlinear control components to tackle  balance control for the VHIP. In particular, we show that capturability can be maintained by restricting the ICI to the input constraint set, which makes our controller succeed in cases beyond those in the state-of-the-art \cite{Caron20b}.

\section{Problem Formulation}

We consider a 2D variable-height inverted pendulum (VHIP) on a horizontal plane, which captures the fundamental features of center of mass (CoM) balancing~\cite{Koolen16,Caron20a}. This model consists of a point mass at the CoM, a ground contact point at the zero moment point (ZMP) \cite{Vukobratovic04}, and a massless prismatic link with a virtual leg stiffness \cite{Farley96} between the CoM and the ZMP. The ground reaction force (GRF) acts along the link, and the ZMP is restricted on the ground surface. The CoM position is denoted by $\c=[c_x,c_z]^T$ with  the equations of motion for the VHIP given by
\begin{equation}\label{vhip}
    \ddot{\c} = \lambda \left( \c - \begin{bmatrix} p \\ 0 \end{bmatrix}\right) - \begin{bmatrix} 0 \\ g\end{bmatrix} 
\end{equation}
where $p$ is the ZMP position, $\lambda>0$ is the virtual leg stiffness and $g$ is the gravitational acceleration. Here, 
the virtual leg stiffness satisfies $\lambda=\frac{f_z}{mc_z}$ with units $\s^{-2}$, where $f_z$ is the vertical GRF and $m$ is the mass. Denote the CoM state as $\x=[\c^T,\dot{\c}^T]^T$ and the input as $\u=[p,\lambda]^T$, 
the state-space VHIP dynamics yields
\begin{equation}\label{vhipx}
	\dot\x = f(\x, \u) = \left[
	\begin{array}{c}
	\dot{\c}\\
	\lambda(\c-\phi(\u))\\
	\end{array}
	\right]
\end{equation}
where $\phi$ is the mapping defined by $\phi([a,b]^T):=[a,\frac{g}{b}]^T$.

To generate a physically realizable trajectory, system (\ref{vhip}) is subject to the following state and input constraints. The state vector is assumed to take values in $\X=\{\x\in \mathbb{R}^4\mid c_z>0\}$. The ZMP is required to lie inside the support polygon \cite{Vukobratovic04}:
\begin{equation}\label{zmp}
    p^-\leq p\leq p^+
\end{equation}
where the pair $(p^-,p^+)$ constitutes the boundary of the support polygon.
On the other hand, the GRF is unilateral and limited, leading to the following constraint:
\begin{equation}\label{uni}
	\lambda^-\leq \lambda\leq \lambda^+
\end{equation}
where $\lambda^->0$.
Accordingly, the overall input constraint set can be written as $\U=\{\u\in \mathbb{R}^2\mid(\ref{zmp})~\&~(\ref{uni})~\mathrm{are}~\mathrm{satisfied}\}$.

For the VHIP, a control policy $\pi(\x) = [\pi_p(\x), \pi_\lambda(\x)]^T$ is a mapping from state to input, where $\pi_p(\x)$ and $\pi_\lambda(\x)$ are referred to as an ankle strategy \cite{Stephens07} and a height variation strategy \cite{Caron20a}, respectively.
Given a control policy $\pi$ and an initial state $\x_0$, the closed-loop state and input trajectories are denoted by $\x(\cdot;\pi,\x_0)$ and $\u(\cdot;\pi,\x_0)$. A control input is called {\em admissible} with respect to a state if the closed-loop trajectories starting from this state satisfy all the constraints. A control policy is called {\em admissible} with respect to a region $\X_0\subset \X$ if 
$\u(\cdot;\pi,\x_0)$ is admissible for all $\x_0 \in \X_0$.

\begin{Definition}[Capture Input/Policy]\label{def_cip}
Given a CoM state $\x_0$, a control input $\u(\cdot)$ is called a \emph{capture input} with respect to $\x_0$ if it makes the VHIP stop at some CoM position $\c^f$, i.e., $\x(t)\rightarrow [(\c^f)^T,\0^T]^T$. Given a region $\X_0\subset \X$, a mapping $\pi:\X_0\rightarrow \mathbb{R}\times(0,\infty)$ is called a \emph{capture policy} with respect to $\X_0$ if $\u(\cdot;\pi,\x_0)$ is a capture input for all $\x_0 \in \X_0$.
\end{Definition}

A capture input/policy will bring the VHIP to a stop, however, it may result in trajectories that violate state or input constraints. This issue is overcome by resorting to the  capturability concept \cite{Koolen12,Posa17}, whose definition can be rephrased in terms of {\em admissible} capture inputs as follows. 

\begin{Definition}[Capturability]
A CoM state $\x_0$ is called \emph{capturable} if there exists an admissible capture input with respect to $\x_0$. The region consisting of all capturable states is called the \emph{capture basin}, denoted by $\Omega_\text{VHIP}$.
\end{Definition}

We now formulate our problems as below.

\begin{Problem}[Capturability Analysis]\label{pro_c}
Given a CoM state $\x_0$, determine whether it is capturable.
\end{Problem}

\begin{Problem}[Balance Control]\label{pro_bl}
Given a target CoM position $\c^d$ and a region $\X_0\subset \X$ of interest, find an admissible capture policy $\pi$ with respect to $\X_0$ such that $\x(t;\pi,\x_0)\to  [(\c^d)^T,\0^T]^T$ for all $\x_0 \in \X_0$. 
\end{Problem}

Capturability analysis aims to explore whether a given CoM state can be paired with an admissible capture input to cease the VHIP. While the balance control problem requires finding an admissible capture policy that not only enables the VHIP to come to a stop, but also needs to regulate the CoM position to the desired one. It is worth to stress that capturability analysis is prerequisite for solving the balance control problem, as any policy solving the balance control problem will necessarily result in trajectories that are composed of capturable states. 

Recall that balancing with the linear inverted
pendulum (LIP) model has been largely solved using the
instantaneous capture point (ICP) \cite{Pratt06}. 
Although the time-varying divergent component of motion (DCM), as a generalization of the ICP to VHIP, has been recently studied in depth \cite{Caron18,Caron20a,Caron20b}, similar analytical results compared to that of the LIP are still missing. This motivates us to present another extension of the ICP, the instantaneous capture input (ICI) to the VHIP in Section \ref{sec_ici}, by concentrating on the properties different from the DCM. As a stepping stone, we further apply this concept to conduct capturability analysis in Section \ref{sec_ca}, as well as to synthesize an admissible capture policy in Section \ref{sec_bc}. Simulations are illustrated in Section \ref{sec_s} to demonstrate our findings.

\section{Instantaneous Capture Input for VHIP}\label{sec_ici}

To better illustrate the idea, we first revisit the ICP concept and its key properties to tackle balancing problems. Then, in light of it, we give a formal definition of the generalized concept and subsequently derive its explicit expression, as well as an intrinsic property relative to the capture input. In comparison, we briefly discuss the difference between the ICI concept and the existing DCM.
\vspace{-10px}
\subsection{The ICP revisited}\label{sec_icp}

Fixing the CoM height to $h$ with constant virtual leg stiffness $\lambda=\lambda_{\text{LIP}}=\frac{g}{h}$, the VHIP \eqref{vhipx} reduces to the celebrated LIP model with fixed CoM height:
\begin{equation}\label{lip}
	\ddot{c}_x=\lambda_\text{LIP}(c_x-p).
\end{equation}
In this case, the system state reduces to a 2D state vector $\x_\text{LIP}=[c_x,\dot{c}_x]^T$, and the input only involves the ZMP position. The analysis and control of the LIP model often relies on the ICP defined as follows:
\begin{equation}\label{icp}
	\xi_\text{ICP}(\x_\text{LIP}):=c_x+\frac{\dot{c}_x}{\sqrt{\lambda_\text{LIP}}}.
\end{equation}
The ICP is a scalar function of the LIP state vector. With the ICP, the dynamics \eqref{lip} can be rewritten as 
\begin{align}
	\dot{\xi}_\text{ICP}=&\sqrt{\lambda_{\text{LIP}}}(\xi_\text{ICP}-p)\label{d_icp}\\
	\dot{c}_x=&\sqrt{\lambda_\text{LIP}}(\xi_\text{ICP}-c_x).\label{d_icp'}
\end{align}
Substituting $p=\xi_\text{ICP}$ into \eqref{d_icp} gives $\dot{\xi}_\text{ICP}\equiv 0$, and the stable pole $-\sqrt{\lambda_\text{LIP}}$ of \eqref{d_icp'} implies $
\x_\text{LIP}(t)\rightarrow [\xi_\text{ICP}(\x_\text{LIP}(0)),0]^T$. This leads to the following key properties for the ICP.

\begin{Property}\label{key_icp}
The ICP defined by \eqref{icp} satisfies
\begin{itemize}
	\item it is a capture policy, i.e., $p=\xi_\text{ICP}(\x_\text{LIP})$ $\Rightarrow$ $\x_\text{LIP}(t)\rightarrow [c_x^f,0]^T$ for some constant $c_x^f$;
	\item its trajectory becomes stationary when it serves as a capture policy, i.e., $p=\xi_\text{ICP}(\x_\text{LIP})$ $\Rightarrow$ $\ddt \xi_\text{ICP}(\x_\text{LIP}(t))\equiv0$.
\end{itemize}
\end{Property}

Recall from Definition \ref{def_cip} that, a capture input for the LIP should make the 2D state $x_\text{LIP}$ converge to $[c_x^f, 0]^T$. It turns out that any capture input can be characterized in terms of the {\em scalar} ICP, as indicated by the next result.

\begin{Proposition}[Capture Input Characterization]\label{ci_lip}
A ZMP signal $p$ is a capture input for the LIP \eqref{lip} if $\xi_\text{ICP}(\x_\text{LIP}(t))$ converges to some constant $\xi^f$. Moreover, $\xi_\text{ICP}(\x_\text{LIP}(t))\rightarrow \xi^f$ $\iff$ $\x_\text{LIP}(t)\rightarrow [\xi^f,0]^T$.
\end{Proposition}

Capturability analysis and balance control of the LIP \eqref{lip} can be handled elegantly using Property \ref{key_icp} and Proposition \ref{ci_lip}, respectively. In view of Property \ref{key_icp}, $p=\xi_\text{ICP}(\x_\text{LIP})$ directly yields the following region of capturable states
\begin{equation}\label{cb_lip}
	\Omega_\text{LIP}=\{\x_\text{LIP}\in \mathbb{R}^2\mid p^-\leq \xi_\text{ICP}(\x_\text{LIP})\leq p^+\}
\end{equation}
which is indeed the capture basin by noticing the effect of \eqref{zmp} on \eqref{d_icp}. Given a target horizontal CoM position $c_x^d\in [p^-,p^+]$, Proposition \ref{ci_lip} implies that to ensure $\x_\text{LIP}(t)\rightarrow [c_x^d,0]^T$, the proposed controller needs to achieve $\xi_\text{ICP}(\x_\text{LIP}(t))\rightarrow \xi_\text{ICP}^d$ with $\xi_\text{ICP}^d=c_x^d$. In view of this, an ICP-based controller with the state-dependent feedback gain $k$ can be constructed as follows
\begin{equation}\label{icp_bc}
	p=\xi_\text{ICP}+k(\xi_\text{ICP}-\xi_\text{ICP}^d)
\end{equation}
which coincides with the one given in \cite{Englsberger15} for static balancing. 
\vspace{-10px}
\subsection{Instantaneous capture input}

Motivated by the importance of the ICP concept in balancing for the LIP, we now extend the ICP to the VHIP model based on the key properties of ICP as stated in Property~\ref{key_icp}. Different from the LIP model, the input of the VHIP model also includes the virtual leg stiffness $\lambda = \frac{f_z}{mc_z}$. The formal definition of the generalization is given as below.

\begin{Definition}[Instantaneous Capture Input]\label{def_ici}
A mapping $\Phi:\X\rightarrow \mathbb{R}\times(0,\infty)$ is said to be an \emph{instantaneous capture input} (ICI) for the VHIP \eqref{vhipx} if 
\begin{itemize}
	\item it is a capture policy, i.e., $\u=\Phi(\x)$ $\Rightarrow$ $\x(t)\rightarrow [(\c^f)^T,\0^T]^T$ for some constant $\c^f$;
	\item its trajectory becomes stationary when it serves as a capture policy, i.e., $\u=\Phi(\x)$ $\Rightarrow$ $\ddt\Phi(\x(t))\equiv\0$.
\end{itemize}
\end{Definition}

We proceed to derive an explicit expression for the ICI notion. Suppose $\xxi=[\xi_p,\xi_\lambda]^T$ satisfies Definition \ref{def_ici}. 
Since $\xxi(\x(t))$ remains constant under the control policy $\u=\xxi(\x)$, the closed-loop system of \eqref{vhipx} can be expressed as
\begin{equation}\label{cls_ici}
	\ddt 
	\left[
	\begin{array}{c}
		\c-\phi(\xxi)\\
		\dot{\c}\\
		\end{array}
		\right]=
			\left[
		\begin{array}{cc}
			\0&I_2\\
			\xi_\lambda I_2&\0\\
		\end{array}
		\right]
		\left[
		\begin{array}{c}
			\c-\phi(\xxi)\\
			\dot{\c}\\
		\end{array}
		\right]
\end{equation}
which is a linear time-invariant system admitting eigenvalues $\pm\sqrt{\xi_\lambda}$. The convergence of $\x$ implies that the initial state $\x_0$ must be an eigenvector corresponding to the stable eigenvalue $-\sqrt{\xi_\lambda}$. Hence, the equation 
\begin{equation}\label{ici_ini}
	\c+\frac{\dot{\c}}{\sqrt{\xi_\lambda(\x)}}=\phi(\xxi(\x))
\end{equation}
holds for $\x=\x_0$. This leads to the upcoming result.

\begin{Theorem}\label{thm1}
The mapping $\xxi=[\xi_p,\xi_\lambda]^T$ defined by
\begin{equation}\label{ici}
\left\{
\begin{array}{lr}
	\xi_p(\x):=c_x+\frac{\dot{c}_x}{\omega} \\
	\xi_\lambda(\x):=\omega^2
\end{array}
\right.
\end{equation}
with $\omega=(\sqrt{\dot{c}_z^2+4c_zg}-\dot{c}_z)/(2c_z)$ solving
\begin{equation}\label{qe}
	c_z\omega^2+\dot{c}_z\omega-g=0
\end{equation}
is the unique ICI for the VHIP \eqref{vhipx}. 
\end{Theorem}

The proof of Theorem \ref{thm1} mainly relies on the following dynamics of $\xxi$ (see Section \ref{dd}):
\begin{align}
\dot{\xxi}=&A(\xxi,\c,\u)(\xxi-\u)\label{d_ici}\\
\dot{\c}=&\sqrt{\xi_\lambda}(\phi(\xxi)-\c)\label{d_ici'}
\end{align}
where 
\begin{equation*}
	A(\xxi,\c,\u)=
	\left[
	\begin{array}{cc}
	\frac{\lambda}{\sqrt{\xi_\lambda}}& \alpha(c_z,\xi_\lambda)(\xi_p-c_x)\\
	0 & \beta(c_z,\xi_\lambda)\\
	\end{array}
	\right]
\end{equation*}
with $\alpha(c_z,\xi_\lambda)=\frac{g}{\sqrt{\xi_\lambda}(c_z\xi_\lambda+g)}$ and $\beta(c_z,\xi_\lambda)=\frac{2c_z\xi_\lambda^{3/2}}{c_z\xi_\lambda+g}$. In fact, by applying $\u=\xxi$, \eqref{d_ici}
gives $\dot{\xxi}\equiv 0$, and $-\sqrt{\xi_\lambda}$ becomes a stable pole of \eqref{d_ici'}, implying that 
$\x(t)\rightarrow [\phi(\xxi(\x_0))^T,\0^T]^T$. The uniqueness immediately follows from the form of \eqref{ici_ini}.

\begin{figure}[tp!] 
	\centering
	\includegraphics[width=0.8\linewidth]{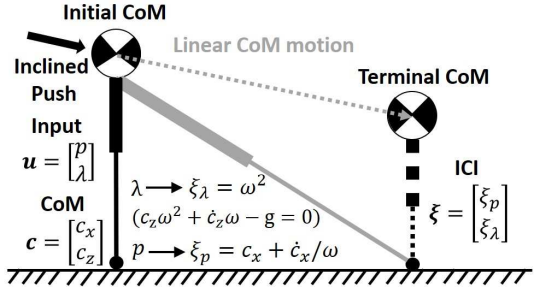}
	\caption{\footnotesize A linear motion of the CoM induced by the proposed ICI. When an inclined push acts on the VHIP, if the control input is instantaneously set to the ICI, then the VHIP moves along the inclined line collinear with the direction of the push, and eventually comes to a stop at the dotted one.}
	\label{ICI}
	\vspace{-15px}
\end{figure}

\begin{Remark}\label{r1}
It is worth mentioning that \eqref{d_ici'} indicates that for any given initial state $\x_0$ of the VHIP, the ICI actually corresponds to a linear motion of the CoM from $\c_0$ to $\phi(\xxi(\x_0))$, as illustrated in Fig. \ref{ICI}.
In addition, from a general (inclined) LIP perspective, any such linear motion generated by our proposed ICI actually corresponds to a LIP model with carefully selected slope and ZMP. Meanwhile, by adopting the VHIP model and proposed ICI concept, the linear motion is automatically generated. Moreover, the proposed ICI provides a compelling route for controller synthesis with the VHIP, enabling trajectories beyond those available via an inclined LIP, as discussed in Section V.
\end{Remark}

We move on to present a characterization of capture inputs via the ICI. Similar to Proposition \ref{ci_lip}, capture inputs of the VHIP can be equivalently defined by the asymptotic behavior of the ICI, provided that $\xi_\lambda$ is lower bounded.
\begin{Proposition}[Capture Input Characterization for VHIP]\label{ci_vhip}
Suppose $\u(\cdot)$ is a control input such that $\xi_\lambda$ has a positive lower bound. Then, $\u(\cdot)$ is a capture input if $\xxi(\x(t))$ converges to some constant vector. Moreover, 
\begin{equation}\label{con}
	\xxi(\x(t))\rightarrow \phi(\c^f)\iff\x(t)\rightarrow
	\left[
	\begin{array}{c}
	\c^f\\
	\0\\
	\end{array}
	\right].
\end{equation}
\end{Proposition}
The proof of Proposition \ref{ci_vhip} is left to the Appendix. It will be seen in Section \ref{sec_bc} that Proposition \ref{ci_vhip} is instrumental for balance control design. Precisely, it uncovers that the regulation of the CoM position is equivalent to that of the ICI towards a constant vector, which is determined by the target CoM position.
\vspace{-10px}
\subsection{Comparison to the DCM}
It is particularly interesting to compare our ICI with the DCM studied in \cite{Hopkins14,Caron20b}, defined by
\begin{equation}\label{dcm}
	\xxi_\text{DCM}:=\c+\frac{\dot{\c}}{\omega_\text{DCM}}
\end{equation}
where $\omega_\text{DCM}$ is the solution of the following Riccati equation
\begin{equation}\label{rie}
	\dot{\omega}_\text{DCM}=\omega_{\text{DCM}}^2-\lambda.
\end{equation}
The name of the DCM comes form the fact that the dynamics of $\xxi_\text{DCM}$, i.e., $\dot{\xxi}_\text{DCM}=\frac{\lambda}{\omega_\text{DCM}}(\xxi_\text{DCM}-\phi(\u))$
represents an unstable mode that is independent of the rest of the state.

The comparison of the ICI and the DCM is as follows. Firstly, the proposed ICI \eqref{ici} and the DCM \eqref{dcm} degenerate to the same ICP \eqref{icp} for the LIP case. In other words, they are both valid generalizations of the ICP to the VHIP. Secondly, these two generalizations concentrate on different aspects of the ICP. Specifically, the DCM aims to decompose the divergent component of the state, while the ICI inherits the role of the ICP as a capture policy, i.e., a kind of control policy that satisfies Definition \ref{def_cip}. 
Thirdly, the ICI depends only on the state, while the DCM dynamics also relies on the auxiliary state $\omega_\text{DCM}$. 
Owing to this reason, our ICI is more tractable for performing capturability analysis as shown in the next section.

\section{Capturability Analysis via ICI}\label{sec_ca}

In this section, we will show how the ICI concept manifests itself in capturability analysis for the VHIP. We first reveal that due to the  properties given in its definition, the ICI naturally induces an inner approximation, i.e., a subset of the capture basin of the VHIP. Furthermore, we derive an outer approximation of the capture basin, i.e., a set contains the capture basin as a subset, through a pair of two-sided approximators of the ICI. The section ends with a discussion on the tightness of the proposed approximations. 
\vspace{-10px}
\subsection{Inner approximation of the capture basin}\label{ca_ia}

According to Definition \ref{def_ici}, when the ICI acts as a control policy, it results in a capture input. Hence, if the CoM state is such that the ICI belongs to the input constraint set, the ICI trajectory will remain stationary and thus the input constraint holds automatically. The next result further ensures the state constraint (the proof is contained in Section \ref{p_p3}).
\begin{Proposition}\label{prop_sc}
Suppose $\u(\cdot)$ is a control input such that $\lambda^-\leq \xi_\lambda\leq \lambda^+$. Then, along the system trajectory starting from any initial state within $\X$, the distance between $c_z$ and the set $H=\{c_z>0\mid \frac{g}{\lambda^+}\leq c_z\leq \frac{g}{\lambda^-}\}$
is decreasing.
\end{Proposition}

Proposition \ref{prop_sc} indicates that if the ICI is always within the input constraint set, then either $c_z$ is always inside $H$ or $c_z$ monotonically converges to $H$, both leading to the satisfaction of the state constraint. As a result, a region of capturable CoM states can be specified directly in terms of the ICI.

\begin{Theorem}[Inner Approximation]\label{inner}
The set 
\begin{equation*}
	\Omega=\{\x\in \X\mid p^-\leq \xi_p(\x) \leq p^+, \lambda^-\leq \xi_\lambda(\x) \leq \lambda^+\}
\end{equation*}
is an inner approximation of the capture basin $\Omega_\text{VHIP}$.
\end{Theorem}

\begin{Remark}\label{r2}
It is worth mentioning that the ICI concept induces linear CoM motions as pointed out in Remark \ref{r1}, and related capturability conditions considering such linear motions have been investigated in the literature \cite{Prete18,Pajon19}. Work in \cite{Prete18} studies multi-contact capturability, and considers general CoM motion along the line in this case. Work in \cite{Pajon19} heuristically fixes a value of $\lambda$ governing motion along a line, with the resulting CoM dynamics similar to those here. However, our ICI concept computes the value of $\lambda$ based on the state of the CoM and induces the above presented capturability condition, which differs fundamentally from the previous results. 
\end{Remark}
\vspace{-10px}
\subsection{Outer approximation of the capture basin}

\begin{figure}[tp!] 
	\centering
	\subfigure[\footnotesize $(c_z,\dot{c}_z)$-slice]{
	\includegraphics[width=0.45\linewidth]{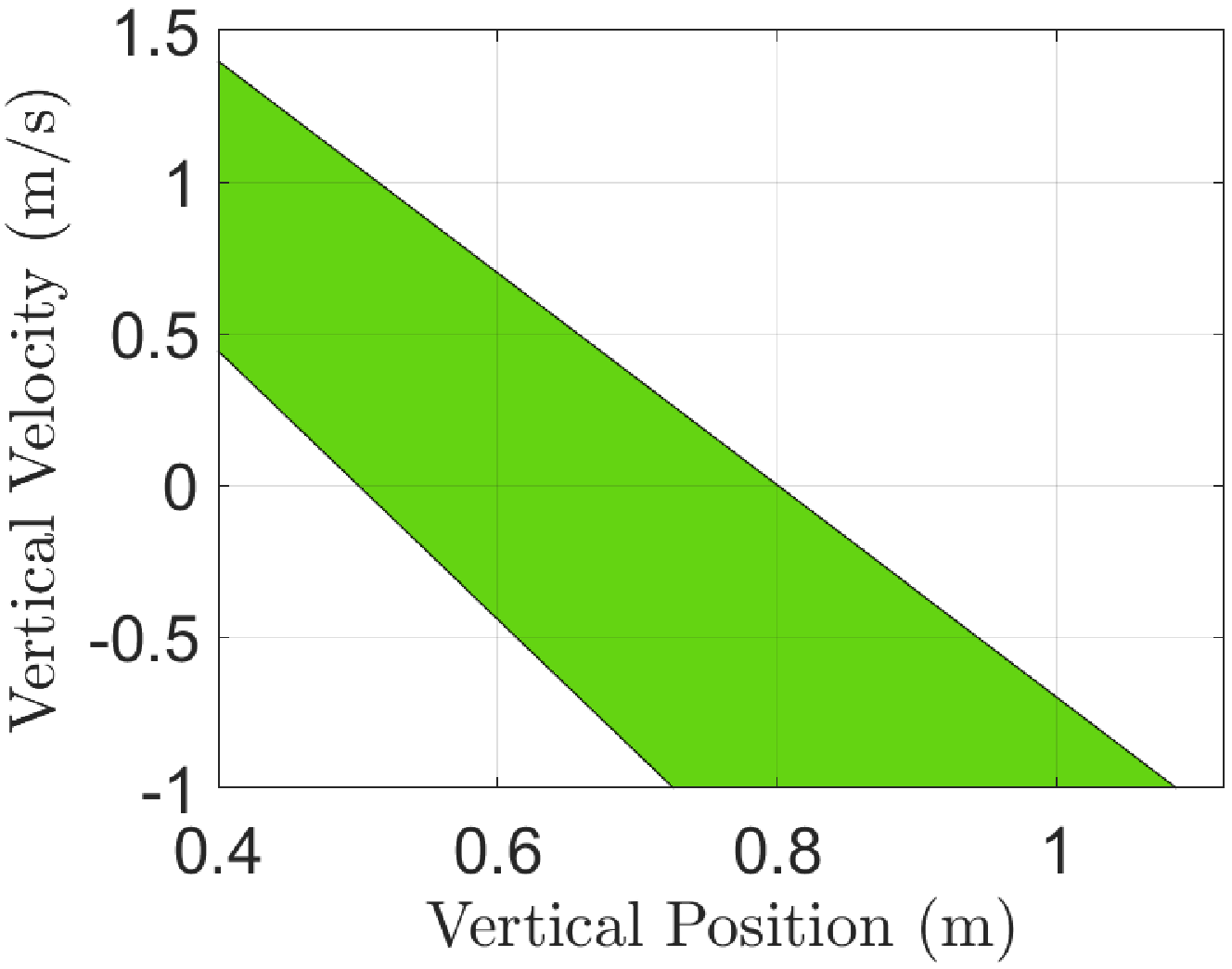}
	\label{sf1}}
	\subfigure[\footnotesize $(c_x,\dot{c}_x)$-slice $(\omega=\sqrt{g/h})$]{
	\includegraphics[width=0.45\linewidth]{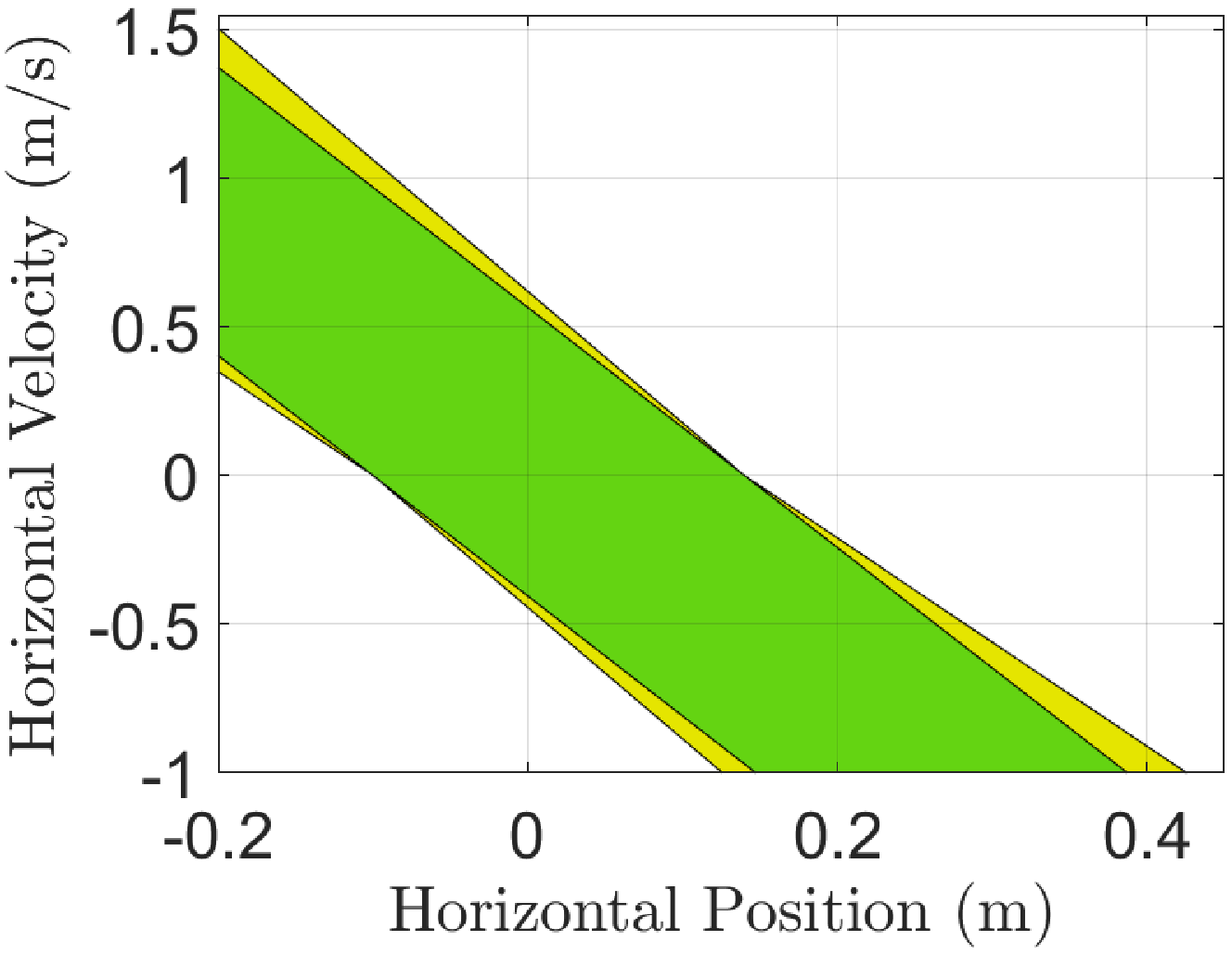}
	\label{sf2}}
	\subfigure[\footnotesize $(c_x,\dot{c}_x)$-slice  $(\omega=\sqrt{\lambda_{\max}})$]{
	\includegraphics[width=0.45\linewidth]{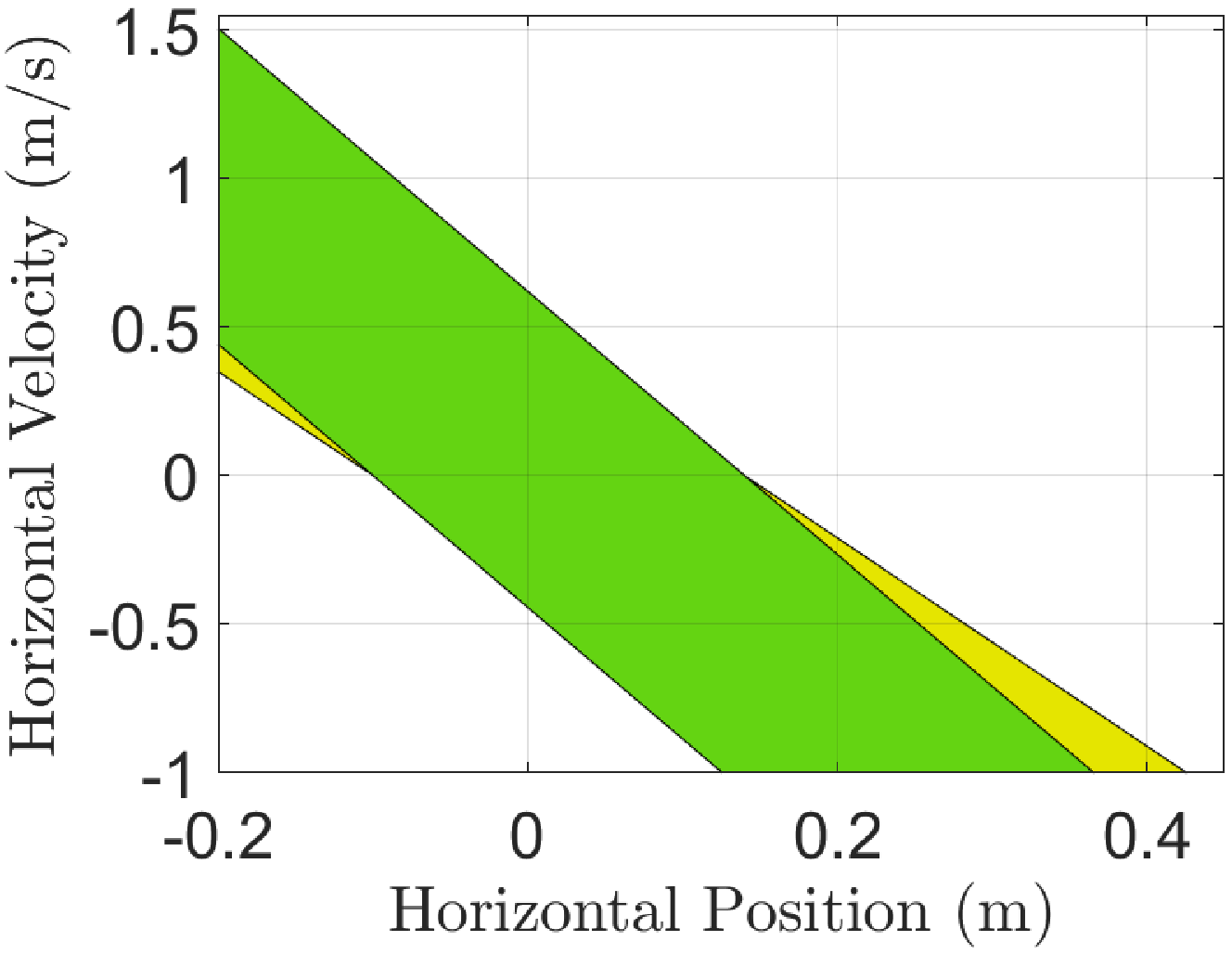}
	\label{sf3}}
	\subfigure[\footnotesize $(c_x,\dot{c}_x)$-slice  $(\omega=\sqrt{\lambda_{\min}})$]{
	\includegraphics[width=0.45\linewidth]{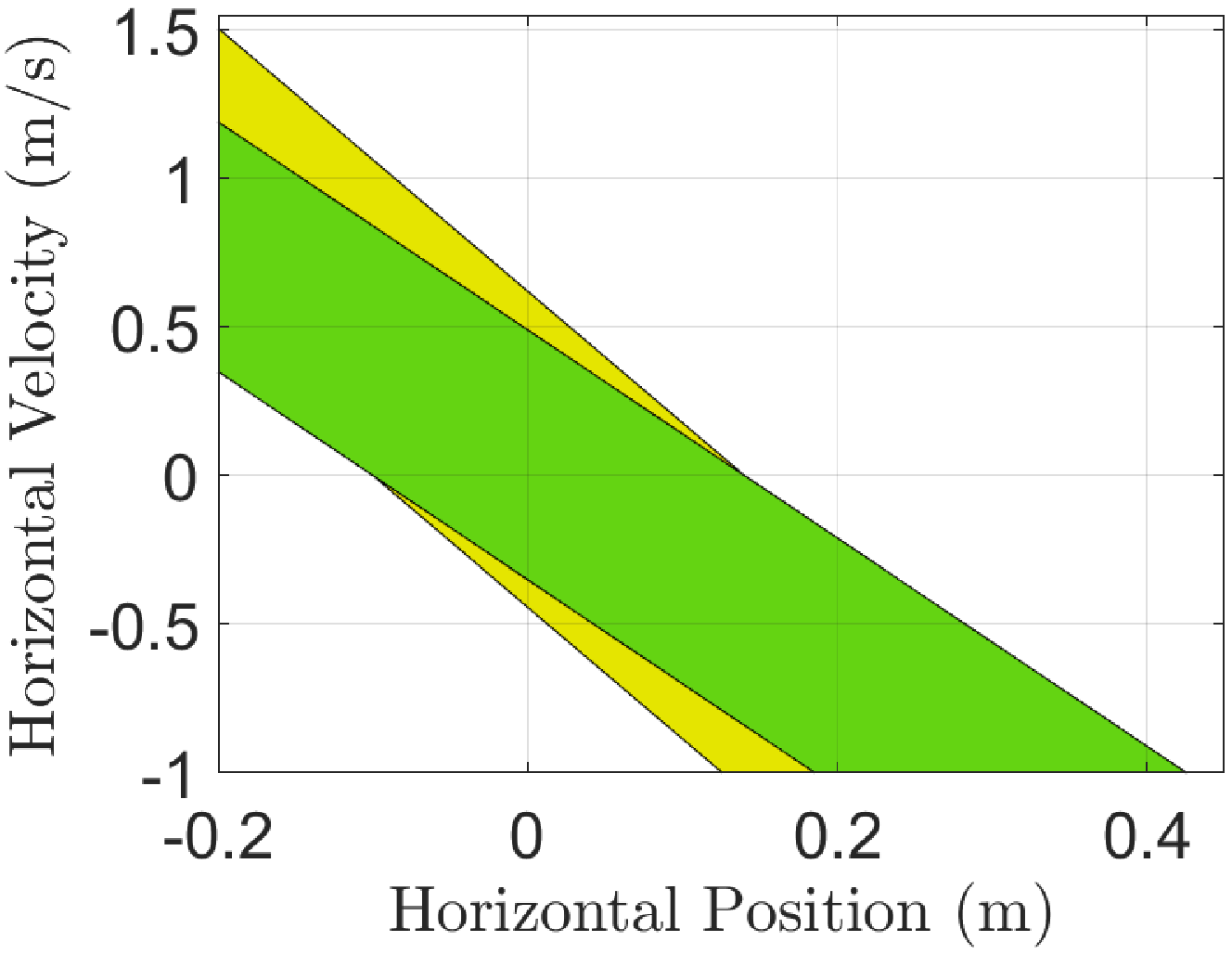}
	\label{sf4}}
	\caption{\footnotesize Visualizations of the inner and outer approximations, using parameters of the UBTECH Walker robot given in Section \ref{sec_s}. The green and the yellow areas represent points corresponding to the inner approximation $\Omega$ and the gap between the inner and the outer approximations $\tilde{\Omega}\setminus \Omega$, respectively.}
	\label{fig3}
	\vspace{-15px}
\end{figure}

Having the inner approximation $\Omega$, we now turn to investigate an outer approximation of the capture basin $\Omega_\text{VHIP}$, or equivalently, the necessity of capturability. Based on Proposition \ref{ci_vhip}, we know that an initial CoM state is capturable only if the corresponding ICI converges to some constant vector. Hence, a natural point of departure for the necessity analysis is to probe into the effect of the input constraint \eqref{zmp}-\eqref{uni} on the dynamics of the ICI \eqref{d_ici} (or \eqref{d_lambda}-\eqref{d_p} in the Appendix). 

To this end, we observe first that the dynamics of $\xi_\lambda$ \eqref{d_lambda} only relies upon the virtual leg stiffness $\lambda$. Since $\beta$ only permits positive values, $\xi_\lambda$ must be inside $[\lambda^-,\lambda^+]$, otherwise $\xi_\lambda$ would monotonically diverge, which contradicts the fact that the ICI is convergent. As a consequence, $\lambda^-\leq \xi_\lambda\leq \lambda^+$ is a necessary condition for capturability.

On the other hand, the argument above can hardly be replicated for the dynamics of $\xi_p$ \eqref{d_p} because of the complexity of the term $\frac{\alpha(c_z,\xi_\lambda)(\xi_\lambda-\lambda)}{\sqrt{\xi_\lambda}}\dot{c}_x$. In view of this, we resort to a pair of two-sided approximators of $\xi_p$ defined by $\xi_p^+:=c_x+\frac{\dot{c}_x}{\sqrt{\lambda^+}}$ and $\xi_p^-:=c_x+\frac{\dot{c}_x}{\sqrt{\lambda^-}}$. Observe that $\min\{\xi_p^+,\xi_p^-\}\leq \xi_p\leq \max\{\xi_p^+,\xi_p^-\}$, and the dynamics of these approximators are
\begin{equation}\label{xi_p'}
    \begin{split}
	\dot{\xi}_p^{+}=&\frac{\lambda}{\sqrt{\lambda^{+}}}(\xi_p^{+}-p)+(1-\frac{\lambda}{\lambda^{+}})\dot{c}_x\\
	\dot{\xi}_p^{-}=&\frac{\lambda}{\sqrt{\lambda^{-}}}(\xi_p^{-}-p)+(1-\frac{\lambda}{\lambda^{-}})\dot{c}_x.
	\end{split}
\end{equation}
It turns out that the form of \eqref{xi_p'} makes the similar argument on $\xi_\lambda$ possible for $\xi_p^+$ and $\xi_p^-$ (see Appendix \ref{p_thm3}). Specifically, the next result shows that it suffices to yield an outer approximation by replacing $p^-\leq \xi_p\leq p^+$ with
\begin{equation}\label{cc_p'}
	\min\{\xi_p^+,\xi_p^-\}\leq p^+~\&~
	\max\{\xi_p^+,\xi_p^-\}\geq p^-.
\end{equation}
\begin{Theorem}[Outer Approximation]\label{outer}
The set
\begin{equation*}
	\tilde{\Omega}=\{\x\in \X\mid
	\eqref{cc_p'}~\mathrm{is}~\mathrm{satisfied}~\&~
	\lambda^-\leq \xi_\lambda\leq \lambda^+\}
\end{equation*}
is an outer approximation of the capture basin $\Omega_\text{VHIP}$.
\end{Theorem}

Theorem \ref{outer} indicates that the inner approximation $\Omega$ is tight with respect to the capture basin $\Omega_\text{VHIP}$. By comparing Theorem \ref{inner} and Theorem \ref{outer}, we know that
the condition $\lambda^-\leq \xi_\lambda\leq \lambda^+$ is valid for all capturable states, and is violated for all non-capturable states.
Since $\xi_\lambda$ is uniquely determined through the quadratic equation \eqref{qe}, we can conclude that the condition provided by $\Omega$ on $(c_z,\dot{c}_z)$ is tight. By tight we do not mean that $\Omega = \Omega_\text{VHIP}$, but rather that these sets share points on their boundary (see Fig. \ref{sf1} for an illustration). 
To visualize the gap between of the inner approximation $\Omega$ and the outer approximation $\tilde{\Omega}$, we also provide examples for the $(c_x,\dot{c}_x)$-slice under three cases (see Fig. \ref{sf2}, \ref{sf3} and  \ref{sf4}). In particular, as illustrated in Fig. \ref{sf2}, under the setting $(c_x,c_z,\dot{c}_z)=(0 \mathrm{m},0.6 \mathrm{m},0 \mathrm{m/s})$, the largest values for $\dot{c}_x$ to lie within $\Omega$ and $\tilde{\Omega}$ are 0.5658 ($\mathrm{m/s}$) and 0.6261 ($\mathrm{m/s}$). Thus, for this case, the gap is at most 0.6261-0.5658=0.0603 ($\mathrm{m/s}$).

\section{ICI-based Balance Control}\label{sec_bc}

In the previous section, we have seen the ICI policy is capable of achieving balancing with height variation, while the terminal CoM position is entirely determined by the initial CoM state. In this section, we further show how to utilize the ICI concept to synthesize feedback controllers so as to regulate the CoM position to a desired one. Our control framework is to convert the balance control problem into the convergence of the ICI dynamics, while preserving capturability and fulfilling the input constraints. For the input constraint handling, we develop an efficient linear programming-based algorithm.
\vspace{-10px}
\subsection{Overview of the control framework}

Let us start by reviewing the tasks of the balance control problem as follows:
1) Convergence: $\x(t)\rightarrow [(\c^d)^T,\0^T]^T$, where $\c^d\in \p=\{\c^d\in \R^2\mid \phi(\c^d)\in \U\}$ is the desired CoM position;
2) Satisfaction of the state constraint: $\x(\cdot)\in \X$;
3) Satisfaction of the input constraint: $\u(\cdot)\in \U$;
4) Capturability maintenance: $\x(\cdot)\in \Omega_\text{VHIP}$.
Note that, Task 4 is not explicitly required for the balance control problem, however, it is necessary since losing capturability would rule out any admissible capture input. In addition, 
Task 2 can be guaranteed by achieving Task 4. In fact, as shown in Theorem \ref{outer}, $\lambda^-\leq \xi_\lambda\leq \lambda^+$ is necessary for capturability, and thus capturability implies $\x(\cdot)\in \X$ on the basis of Proposition \ref{prop_sc}.

Based on Proposition \ref{ci_vhip}, it suffices to ensure Task 1 via $\xxi(\x(t))\rightarrow \xxi^d=[\xi_p^d,\xi_\lambda^d]^T$,
where $\xxi^d=\phi(\c^d)$ is the desired ICI vector generated by the desired CoM position. 
Intuitively, analogous to the ICP-based controller \eqref{icp_bc}, one may expect to use a controller of the form 
\begin{equation}\label{cl'}
	\u=\xxi+K(\xxi-\xxi^d)
\end{equation}
where $K=\mathrm{diag}(k_1,k_2)$ is a diagonal matrix with feedback gains $k_1$ and $k_2$ as functions of the CoM state $\x$.
However, the presence of the upper-right element of the matrix $A$ in the dynamics of the ICI \eqref{d_ici} 
makes the controller \eqref{cl'} unsuitable for realizing Task 1 and Task 4 simultaneously.

\begin{figure}[t!]
	\centering
	\includegraphics[width=0.8\linewidth]{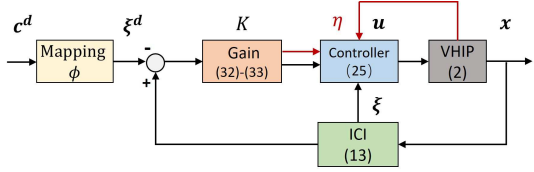}
	\caption{Block diagram of the ICI-based control framework. The red lines with arrows represent
	extra feedback of the controller \eqref{cl} compared to the naive one \eqref{cl'}, i.e., the composition of 
	the nonlinear compensation term $\eeta$.}
	\label{fig_bd}
	\vspace{-15px}
\end{figure}

To eliminate such nonlinearity, we add an nonlinear compensation term $\eeta$ to the controller \eqref{cl'}, which is defined by
\begin{equation}\label{ct}
	\eeta:=
	\left[
	\begin{array}{c}
	\eta_p\\
	0\\
	\end{array}
	\right]
	=\left[
	\begin{array}{c}
	-\frac{k_2\alpha(c_z,\xi_\lambda)(\xi_\lambda-\xi_\lambda^d)}{\xi_\lambda+k_2(\xi_\lambda-\xi_\lambda^d)}\dot{c}_x\\
	0\\
	\end{array}
	\right]
\end{equation}
This leads to the controller of the form
\begin{equation}\label{cl}
	\u=\xxi+K(\xxi-\xxi^d)+\eeta
\end{equation}
which consists of a feedforward $\xxi$, an error feedback with feedback gains $K(\xxi-\xxi^d)$, and a nonlinear compensation $\eeta$. The block diagram of \eqref{cl} is illustrated in Fig. \ref{fig_bd}.
It will be seen in Theorem \ref{lem1} that the controller \eqref{cl} is superior to fulfill both Task 1 and Task 4 under some mild conditions. Hence, the problem reduces to designing an optimization algorithm that enforces the feedback gains to fulfill the conditions in Theorem \ref{lem1} as well as the input constraint.

\vspace{-10px}
\subsection{Convergence with capturability maintenance}

The reason for the form of the controller \eqref{cl} is that it results in the following closed-loop system
\begin{equation}\label{cls}
	\dot{\xxi}=-K\Lambda(\x,\u)(\xxi-\xxi^d)
\end{equation}
where $\Lambda(\x,\u)=\mathrm{diag}(\frac{\lambda}{\sqrt{\xi_\lambda}},\beta(c_z,\xi_\lambda))$ is composed
of exactly the diagonal elements of the matrix $A$ in \eqref{d_ici}. The next result shows that the controller \eqref{cl} has the advantage that, convergence  can be easily achieved while preserving capturability under some mild conditions.
\begin{Theorem}\label{lem1}
Consider the closed-loop system \eqref{vhipx} under the controller \eqref{cl}, where the feedback gains are functions of the CoM state.
Suppose that the initial state $\x_0\in \Omega_\text{VHIP}$, the desired CoM position $\c^d\in \p$ and the feedback gains $k_i\geq \epsilon$, $i=1,2$ with $\epsilon>0$ are such that $\lambda$ has a positive lower bound.
Then, $\xxi(\x(t))\rightarrow \phi(\c^d)$ and $\x(\cdot)\in \Omega_\text{VHIP}$.
\end{Theorem}

The proof of Theorem \ref{lem1} mainly follows from the fact that under the conditions in Theorem \ref{lem1}, the closed-loop system \eqref{cls} admits a quadratic Lyapunov function $V(\xxi)=V_1(\xi_p)+V_2(\xi_\lambda)=\frac{1}{2}(\xi_p-\xi_p^d)^2+\frac{1}{2}(\xi_\lambda-\xi_\lambda^d)^2$, and moreover, $V_1$ and $V_2$ are strictly decreasing along the closed-loop system trajectory. This is the reason that the controller \eqref{cl} with the additional nonlinear compensation term $\eeta$ is capable of retaining capturability, reminiscent of the ICP-based controller \eqref{icp_bc} as discussed in Section \ref{sec_icp}.
\vspace{-10px}
\subsection{Linear programming-based feedback gains algorithm}

We turn to investigate the optimal search for suitable feedback gains that satisfy the conditions in Theorem \ref{lem1} as well as the input constraint \eqref{zmp}-\eqref{uni}. Under the controller \eqref{cl}, the input constraint reduces to
\begin{align}
	\lambda^--\xi_\lambda\leq& k_2(\xi_\lambda-\xi_\lambda^d)\leq \lambda^+-\xi_\lambda\label{ic1}\\
	p^--\xi_p\leq& k_1(\xi_p-\xi_p^d)+\eta_p(k_2,\x,\xi_\lambda^d)\leq p^+-\xi_p.\label{ic2}
\end{align}
A naive approach is to consider the optimization problem
\begin{equation}\label{non-convex}
	\begin{split}
	&\min_{k_1,k_2}~-k_1-c k_2\\
	&~\mathrm{s.t.}~~\eqref{ic1},\eqref{ic2},~\epsilon \leq k_i\leq M,~i=1,2\\
	\end{split}
\end{equation}
where $c>0$ is the weight parameter, $\epsilon$ and $M$ with $0<\epsilon\ll M$ are minimum and maximum allowable values of the feedback gains chosen by the user. Nevertheless, a direct obstruction is due to the relation between $\eta_p(k_2,\x,\xi_\lambda^d)$ and $k_2$, which renders \eqref{non-convex} as a non-convex optimization problem. 

To handle the non-convex constraint \eqref{ic2}, we adopt the idea of convex approximation: bounding $\eta_p$ by means of
\begin{equation}\label{ic1'}
	\gamma(p^--\xi_p)\leq \eta_p(k_2,\x,\xi_\lambda^d)\leq \gamma(p^+-\xi_p)
\end{equation}
where $\gamma\in (0,1)$ is a user-specified parameter determining the influence of $\eta_p$ on \eqref{ic2}. Given the current state $\x$, the constraint \eqref{ic1'} is indeed linear with respect to $k_2$. This can be seen by multiplying all sides of \eqref{ic1'} by the denominator of $\eta_p$ as indicated by \eqref{ct}, which gives
\begin{equation}\label{ic1''}
	\begin{split}
		-(\xi_\lambda-\xi_\lambda^d)[\alpha\dot{c}_x+\gamma(p^+-\xi_p)]k_2\leq& \gamma(p^+-\xi_p)\xi_\lambda\\
		(\xi_\lambda-\xi_\lambda^d)[\alpha\dot{c}_x+\gamma(p^--\xi_p)]k_2\leq& -\gamma(p^--\xi_p)\xi_\lambda.
	\end{split}
\end{equation}
After specifying the feedback gain $k_2$, the constraint \eqref{ic2} for the remaining feedback gain $k_1$ becomes linear. As a consequence, the non-convex optimization \eqref{non-convex} is converted to the following linear programming (LP) problems in series:
\begin{equation}\label{lp1}
		\begin{split}
			&\min_{k_2}~-k_2\\
			&~\mathrm{s.t.}~~\eqref{ic1}, \eqref{ic1''},~
			\epsilon\leq k_2\leq M
		\end{split}
	\end{equation}
and subsequently
	\begin{equation}\label{lp2}
		\begin{split}
			&\min_{k_1}~-k_1\\
			&~\mathrm{s.t.}~~\eqref{ic2},~
			\epsilon\leq k_1\leq M.
		\end{split}
\end{equation}

Our main result for balance control is summarized as below (see Appendix for the proof).

\begin{Theorem}\label{thm_bc}
Given any initial CoM state $\x_0\in\mathrm{int}(\Omega)$, i.e., the interior of $\Omega$, and any desired CoM position $\c^d\in \p$, the controller of the form (\ref{cl}) with the feedback gains solving the LP problems \eqref{lp1}-\eqref{lp2} under sufficiently small $\epsilon>0$ addresses the balance control problem for the VHIP \eqref{vhipx}, i.e., $\x(t)\rightarrow [(\c^d)^T,\0^T]^T$, $\x(\cdot)\in \X$ and $\u(\cdot)\in \U$.
\end{Theorem}

The underlying philosophy of our controller differs from the DCM-based approach presented in 
\cite{Caron20b}. In particular, the DCM-based approach requires a linear approximation of the closed-loop system to coincide with the prescribed closed-loop system. In contrast, the closed-loop system under our nonlinear feedback controller \eqref{cl} takes the exact form of \eqref{cls}, enabling us to handle cases for which the DCM-based approach would fail to solve, as illustrated in the next section.

\section{Simulation}\label{sec_s}

\begin{figure*}[tp!]
	\centering
	\subfigure[\footnotesize DCM-based result ($\omega_{\text{DCM}}(0)=\sqrt{g/h}$)]{%
		\includegraphics[width=0.3\linewidth]{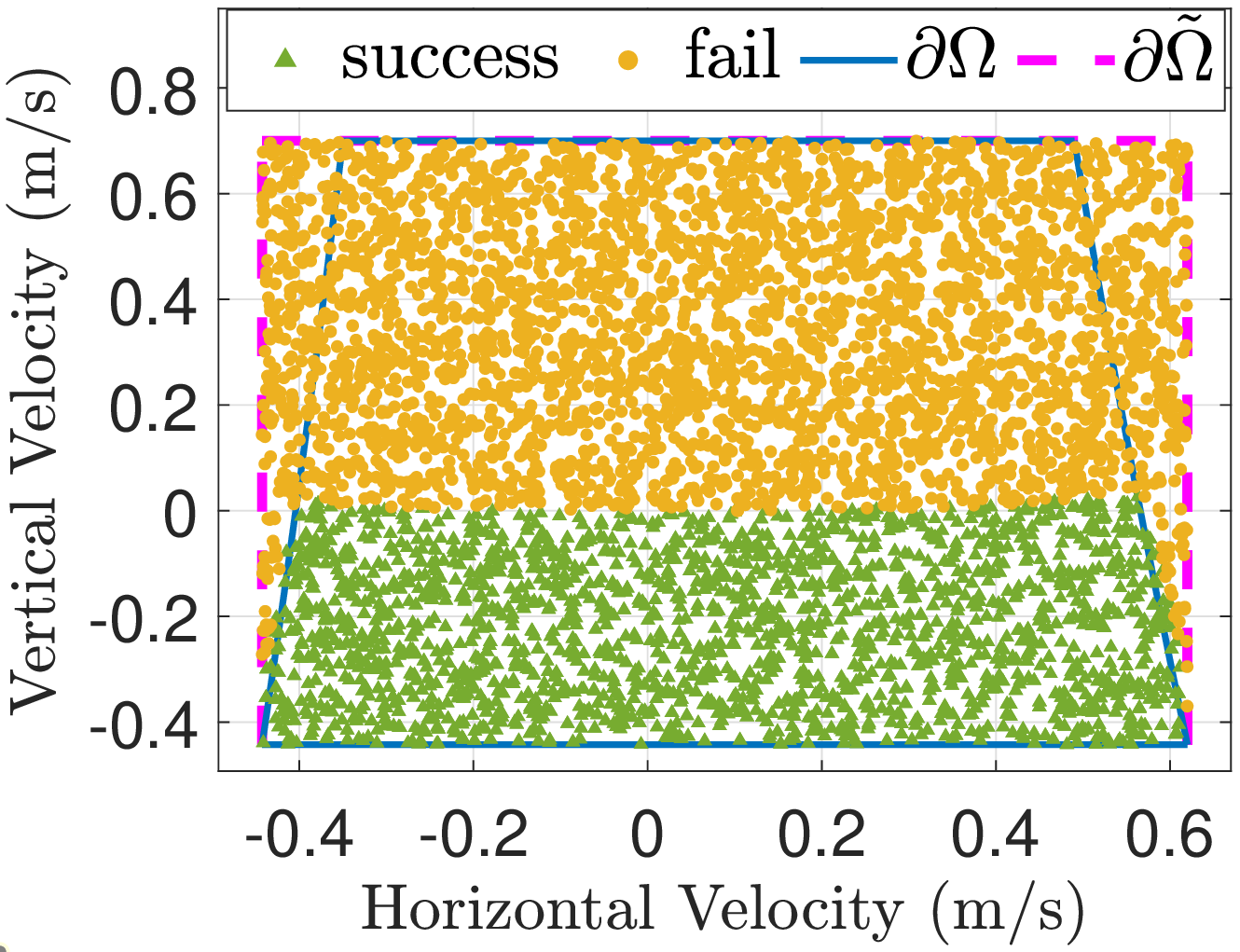}
		\label{fig4a}}
	\subfigure[\footnotesize DCM-based result ($\omega_{\text{DCM}}(0)=3.6$)]{%
		\includegraphics[width=0.3\linewidth]{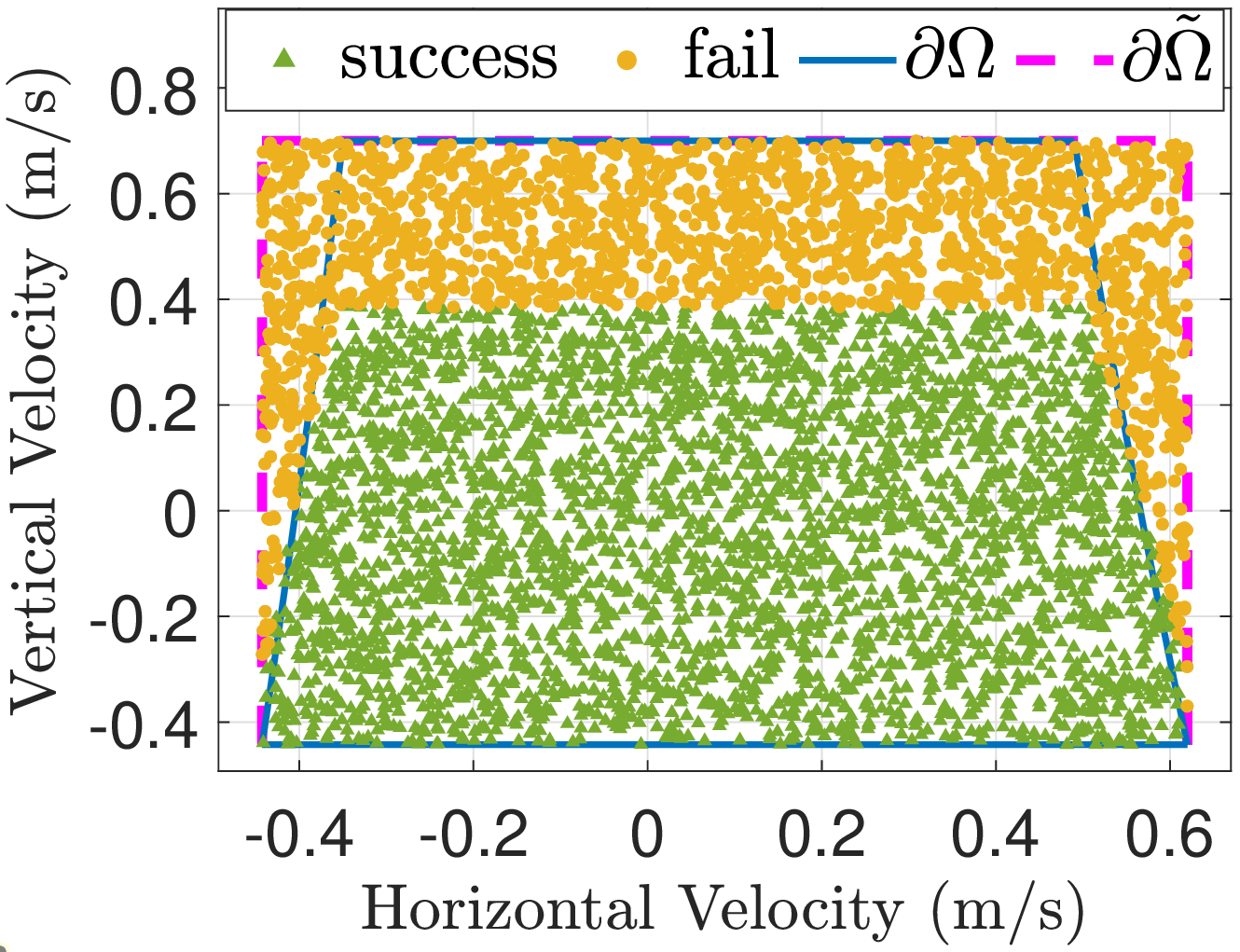}
		\label{fig4b}}
	\subfigure[\footnotesize ICI-based result]{%
		\includegraphics[width=0.3\linewidth]{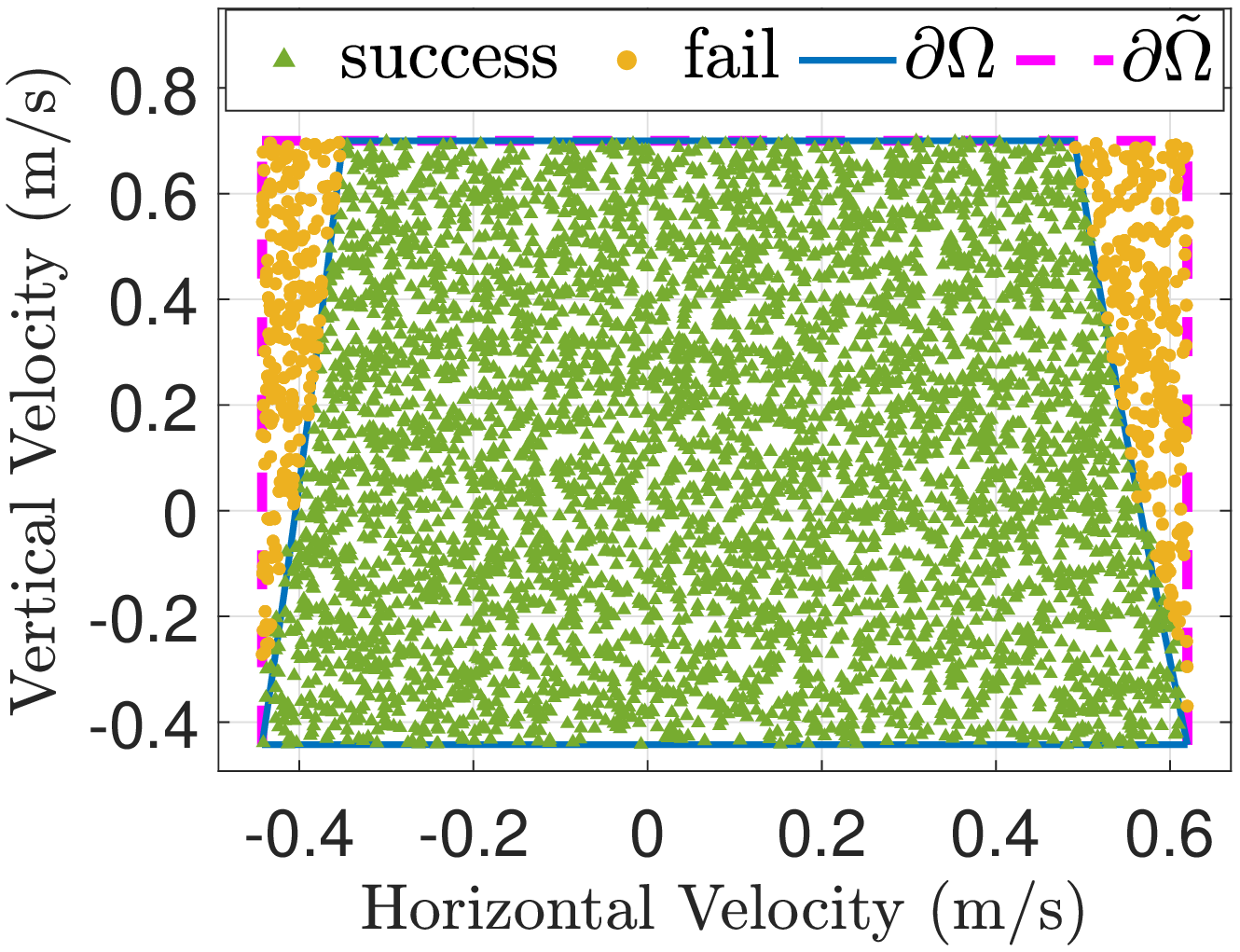}
		\label{fig4c}}
	\caption{\footnotesize Monte Carlo simulation results of the ICI-based nonlinear controller and the DCM-based linearization controller ($10^4$ samples).}
	\label{fig4}
	\vspace{-15px}
\end{figure*}

We validate the ICI-based controller \eqref{cl} solving the LP problems \eqref{lp1}-\eqref{lp2} using MATLAB simulation. 
The parameters of the VHIP model stems from the physical properties of the UBTECH Walker robot. The mass and foot size of Walker are about $70\mathrm{kg}$ and $0.24 \mathrm{m}$, and the range of its CoM height is between $0.5 \mathrm{m}$ and $0.8 \mathrm{m}$. Based on these properties, the parameters are set to $p^-/p^+=-0.1/0.14 (\mathrm{m})$,
$\lambda^-=9.8/0.8=12.25 (\mathrm{s}^{-2})$ and 
$\lambda^+=9.8/0.5=19.6 (\mathrm{s}^{-2})$.
In simulation, we assume the VHIP encounters sudden pushes, which results in impulses on the CoM. Since an impulse depends linearly on the change of velocity, the effect of the pushes can be modelled as instantaneous velocity changes of the CoM.
The task is to synthesize ankle and height variation strategies to achieve regulation of the CoM position toward the prescribed one. 
In comparison, we also demonstrate the performance of the DCM-based linearization controller \cite{Caron20b} as well as the ICP-based controller \eqref{icp_bc} \cite{Englsberger15}.
\vspace{-10px}
\subsection{ICI-based controller versus DCM-based controller}

In this comparison, we perform Monte Carlo simulations under different initial velocities, as illustrated in Fig. \ref{fig4}. Fixing the nominal CoM position as $[0 \mathrm{m}, 0.6\mathrm{m}]^T$, $10^4$ initial CoM velocities are randomly generated within the outer approximation $\tilde{\Omega}$.
The target CoM position is chosen to be the same as the nominal one. The experiments are carried out with the terminal time $t_f=4\mathrm{s}$, and the success criterion is to meet $\max\{||\c(t_f)-\c^d||,||\dot{\c}(t_f)||\}<0.01$.
The controller parameters are selected as $\epsilon=10^{-3}$, $M=10$ and $\gamma=0.1$ for the ICI-based approach, and $\epsilon=10^{-3}$, $k=10$, $\omega_{\text{DCM}}(0)=\sqrt{g/h}$ as well as the reference $\xi_{\text{DCM}}^d=\c^d$, $\omega_{\text{DCM}}^d=\sqrt{g/c_z^d}$ for the DCM-based approach. 
Fig. \ref{fig4a} and Fig. \ref{fig4c} show that two controllers perform similarly for negative vertical velocities, and the DCM-based controller is capable of handling some edge cases that fail for the ICI-based controller.
However, for other initial CoM velocities, the DCM-based controller fails in large regions where the ICI-based controller succeeds. Note that the DCM-based approach heavily relies on the choice of $\omega_{\text{DCM}}(0)$. In fact, if $\omega_{\text{DCM}}(0)$ is switched to $3.6\in [\sqrt{\lambda^-},\sqrt{\lambda^+}]~ (\mathrm{s^{-2}})$, the DCM-based performance improves significantly as illustrated in Fig. \ref{fig4b}. In contrast, the ICI-based controller needs no additional parameter tuning, and guarantees that the domain of attraction at least contains the interior of $\Omega$.

\vspace{-10px}
\subsection{ICI-based controller versus ICP-based controller}

\begin{figure}[tp!]
	\centering
	\subfigure[\footnotesize ICP-based Position]{%
		\includegraphics[width=0.48\linewidth]{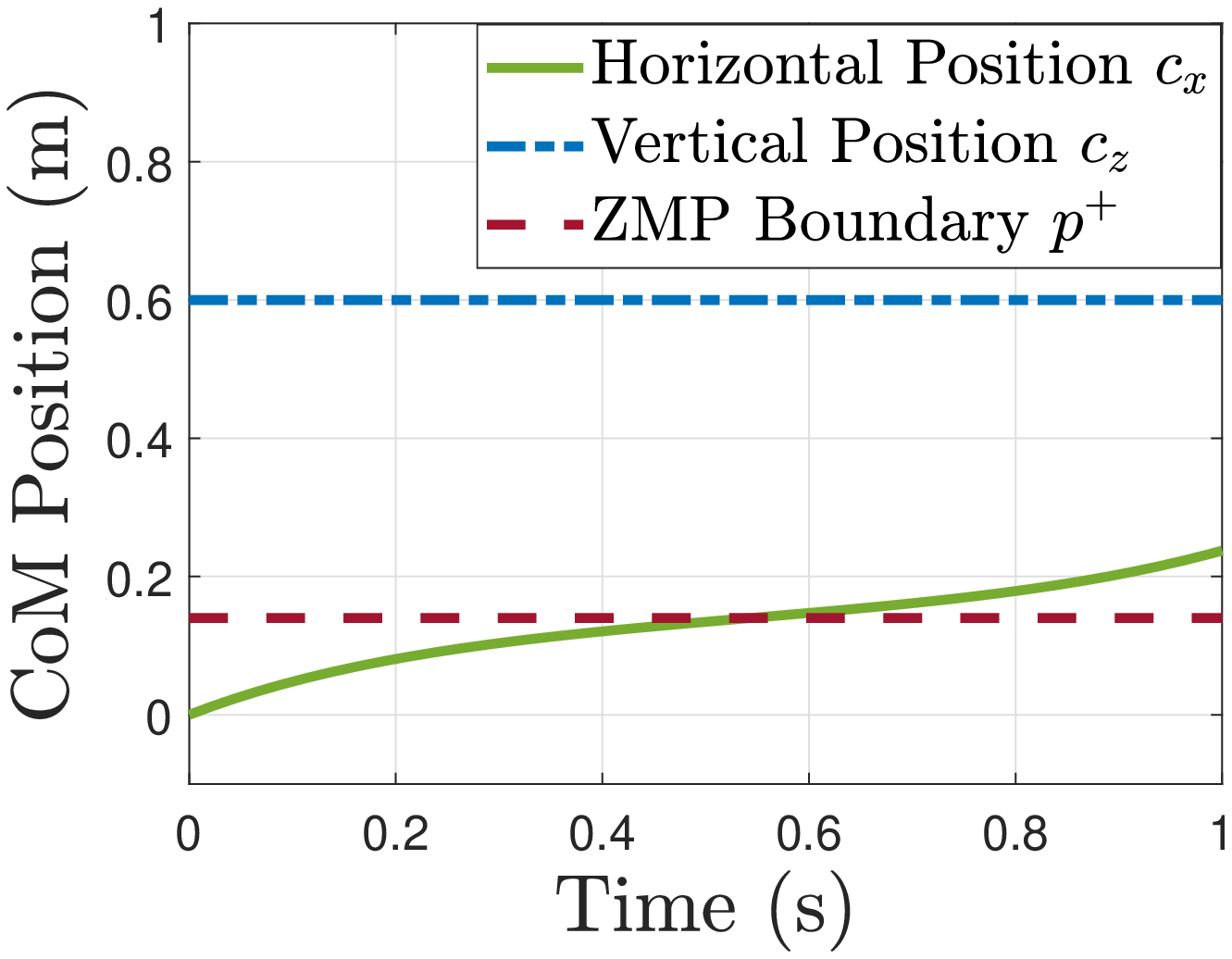}
		\label{fig5_a}}
	\subfigure[\footnotesize ICI-based Position]{%
		\includegraphics[width=0.48\linewidth]{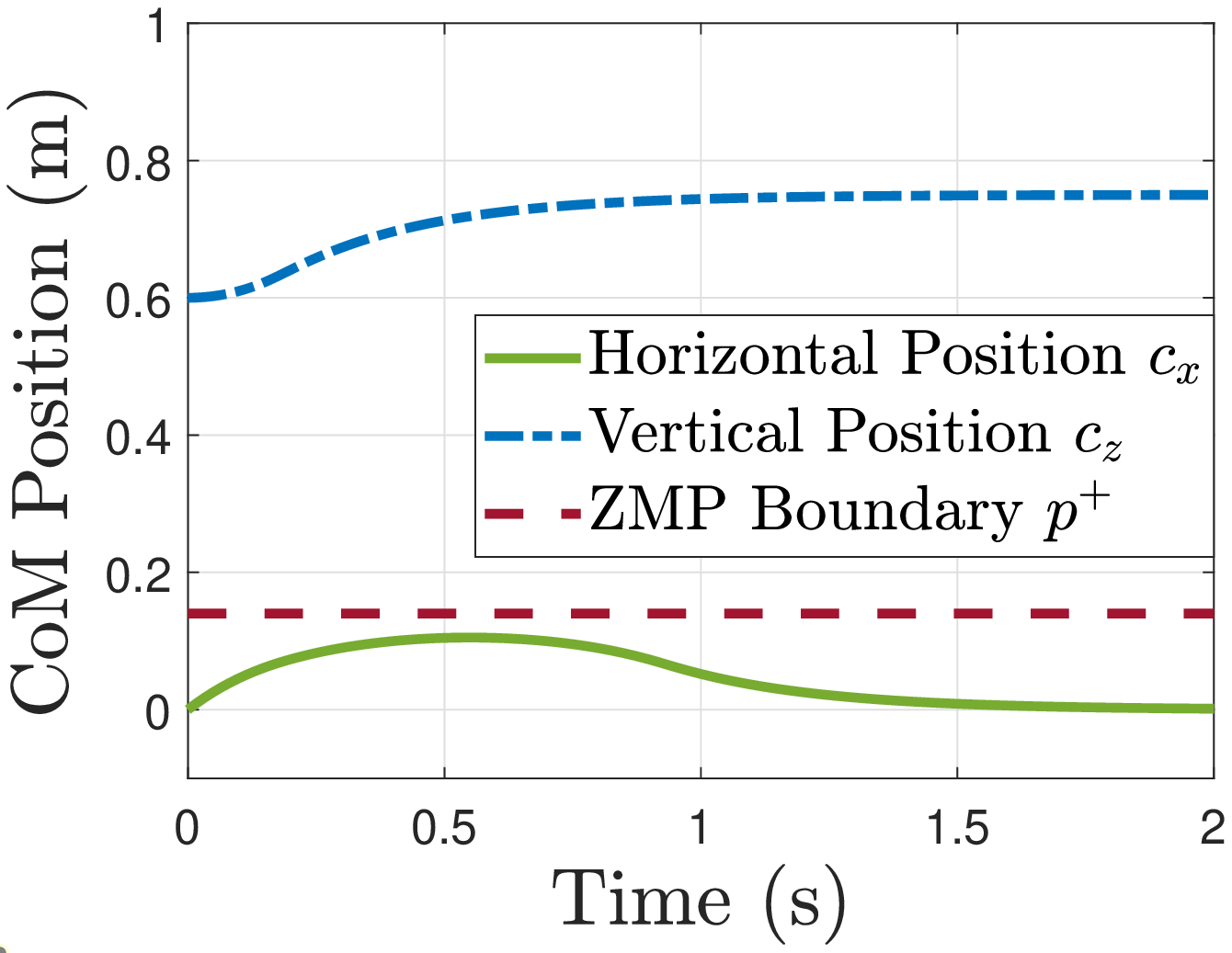}
		\label{fig5_b}}
	\subfigure[\footnotesize ICI-based Input]{%
		\includegraphics[width=0.48\linewidth]{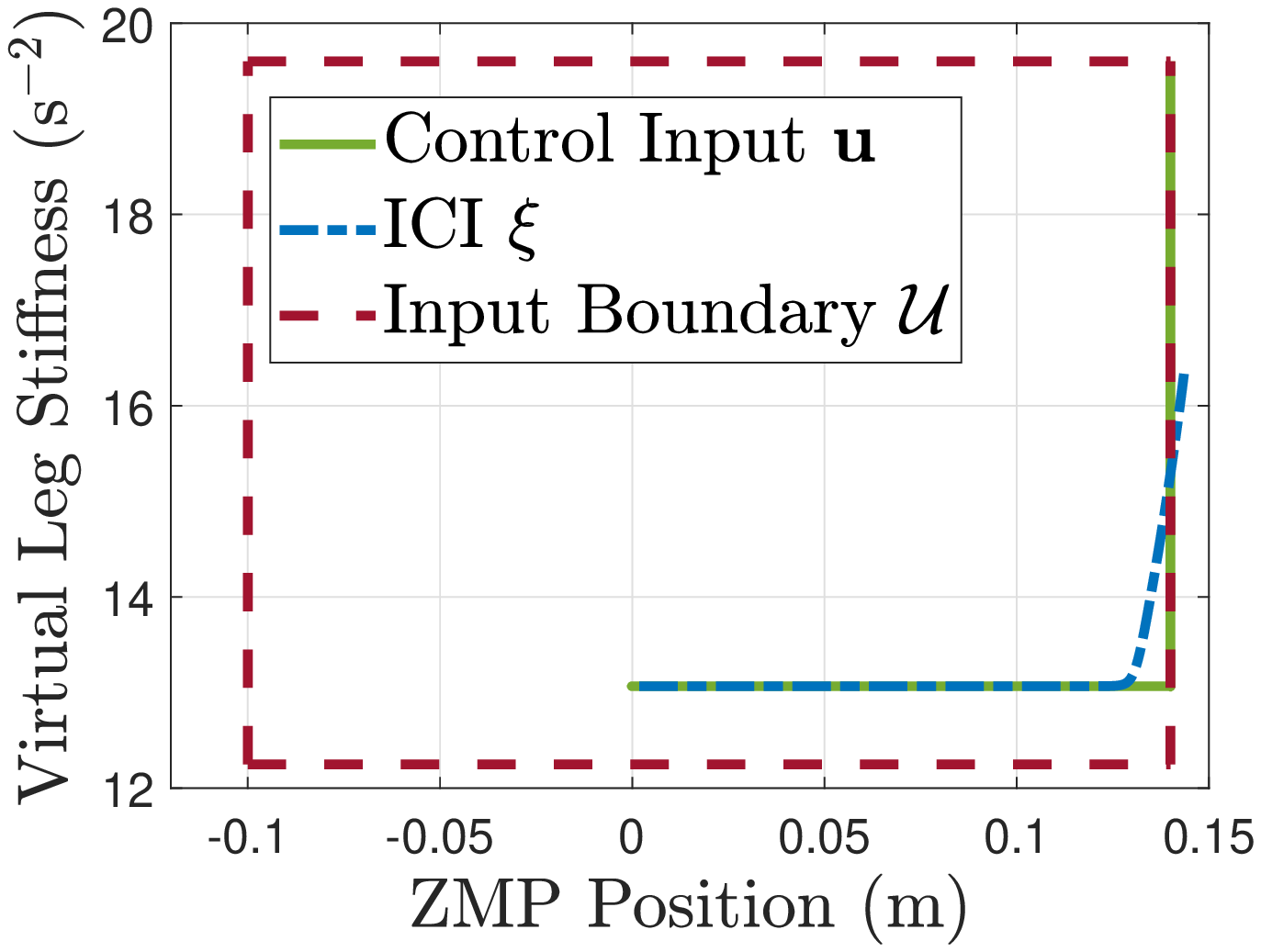}
		\label{fig5_c}}
	\subfigure[\footnotesize ICI-based Feedback Gains]{%
		\includegraphics[width=0.48\linewidth]{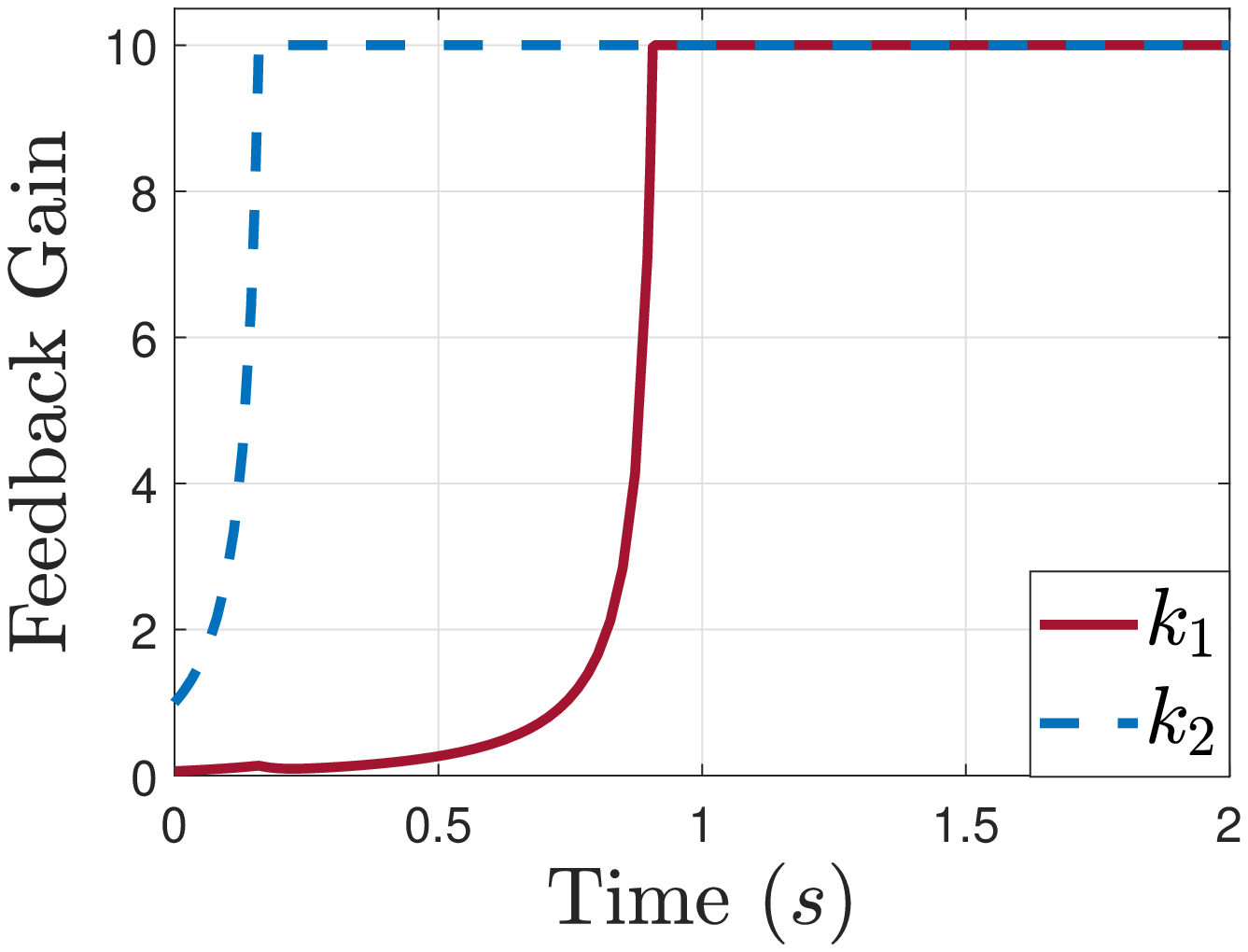}
		\label{fig5_d}}
	\caption{\footnotesize Simulation results of the ICI-based nonlinear controller and the ICP-based linear controller for the case when the initial CoM state is outside $\Omega$.}
	\label{fig5}
	\vspace{-15px}
\end{figure}

Next, a comparison to the ICP-based controller is made under a scenario when the initial CoM state is outside $\Omega$.
The nominal CoM position is still set to be $[0 \mathrm{m}, 0.6\mathrm{m}]^T$, and the change of velocity resulting from a horizontal impulse is $[0.58 \mathrm{m/s}, 0\mathrm{m/s}]^T$.
Now, the corresponding CoM state is outside the inner approximation $\Omega$ since $\xi_p(\x_0)=0.1435>0.14$. As discussed in Section \ref{ca_ia}, the ICP-based controller cannot handle this case via fixing the CoM height, which is demonstrated in Fig. \ref{fig5_a}. Nevertheless, if we set the target CoM position as $\c^d=[0 \mathrm{m}, 0.75 \mathrm{m}]^T$, and apply the ICI-based controller, balancing can be achieved as shown in Fig. \ref{fig5_b}. In particular, it can be seen in Fig. \ref{fig5_c} that, although the ICI at the initial CoM state does not belong to the input constraint set, it eventually enters the input constraint set and then remain there indefinitely. The feedback gains obtained by solving  \eqref{lp1}-\eqref{lp2} are shown in Fig. \ref{fig5_d}.

\section{Conclusion}

We have introduced a novel concept called the ICI, which has been shown to be a generalization of the ICP to the VHIP model based on its properties as a control policy. This characterization directly enables a tight closed-form inner approximation of the capture basin. Moreover, for those CoM states strictly within the inner approximation, we have shown that balance control can be resolved effectively via an ICI-based controller with a linear programming-based feedback gains algorithm. As a byproduct, we provide some insight for capturability maintenance, that is, to confine the ICI trajectory within the prescribed input constraint set. This constraint extends the longstanding stability criterion for the LIP model that requires the ICP to be with the support polygon.

\appendix

\subsection{Derivation of the dynamics \eqref{d_ici}-\eqref{d_ici'}}\label{dd}

It remains to show $\xxi$ and $\c$ evolve as 
\eqref{d_ici}-\eqref{d_ici'}. The dynamics of \eqref{d_ici'} follows from \eqref{ici} and \eqref{qe}. Differentiating both sides of \eqref{qe}, together with substituting the second equation of \eqref{ici} gives
$\dot{\xi}_\lambda=-[\dot{c}_z\xi_\lambda+(\lambda c_z-g)\sqrt{\xi_\lambda}]/(c_z+\frac{\dot{c}_z}{2\sqrt{\xi_\lambda}})$. By invoking \eqref{qe}, $\dot{\xi}_\lambda$ can be expressed compactly as 
\begin{equation}\label{d_lambda}
	\dot{\xi}_\lambda=\beta(c_z,\xi_\lambda)(\xi_\lambda-\lambda).
\end{equation}
Additionally, differentiating both sides of the first equation of \eqref{ici} yields $\dot{\xi}_p=\frac{\lambda}{\sqrt{\xi_\lambda}}[(\xi_p-p)+\dot{c}_x(\frac{\sqrt{\xi_\lambda}}{\lambda}-\frac{1}{\sqrt{\xi_\lambda}}-\frac{\dot{\xi}_\lambda}{2\lambda\xi_\lambda})]$. Owing to \eqref{qe}, the derivative of $\xi_p$ can be further simplified as
\begin{equation}\label{d_p}
	\dot{\xi}_p=\frac{\lambda}{\sqrt{\xi_\lambda}}(\xi_p-p)+\frac{\alpha(c_z,\xi_\lambda)(\xi_\lambda-\lambda)}{\sqrt{\xi_\lambda}}\dot{c}_x.
\end{equation}
Combining \eqref{d_lambda} and \eqref{d_p} yields \eqref{d_ici} due to the form of \eqref{d_ici'}.
\vspace{-10px}
\subsection{Proof of Proposition \ref{ci_vhip}}

If $\u(\cdot)$ is a capture input, i.e., the right-hand side of \eqref{con} holds for some $\c^f$, then \eqref{d_ici'} yields that $\phi(\xxi)\rightarrow \c^f$, and thus the left-hand side of \eqref{con} is valid due to the fact that $\phi^2$ is an identity mapping. Conversely, suppose $\u(\cdot)$ makes $\xxi\rightarrow \phi(\c^f)$ for some $\c^f$. 
Note that, \eqref{d_ici'} can be rewritten as
$\dot{\c}=-\sqrt{\xi_\lambda}(\c-\c^f)+\sqrt{\xi_\lambda}(\phi(\xxi)-\c^f)$.
Then $\xxi-\phi(\c^f)\rightarrow \0$ implies that $\phi(\xxi)-\c^f\rightarrow \0$, which leads to $\c\rightarrow\c^f$ and $\dot{\c}\rightarrow \0$ since 
$\xi_\lambda$ has a positive lower bound.
\vspace{-10px}
\subsection{Proof of Proposition \ref{prop_sc}}\label{p_p3}

The dynamics \eqref{d_ici'} indicates that $\dot{c}_z=\sqrt{\xi_\lambda}(\frac{g}{\xi_\lambda}-c_z)$. Hence, if $c_z>\frac{g}{\lambda^-}$, then $\dot{c}_z<\sqrt{\xi_\lambda}(\frac{g}{\xi_\lambda}-\frac{g}{\lambda^-})\leq0$, implying that $c_z$ must be strictly decreasing. Similar arguments applied to the case of $c_z<\frac{g}{\lambda^+}$ yields that $c_z$ must be strictly increasing.
\vspace{-10px}
\subsection{Proof of Theorem \ref{outer}}\label{p_thm3}

It remains to show the necessity of \eqref{cc_p'}, which is conducted by contradiction. Suppose a capturable state with $\dot{c}_x>0$ satisfies $\xi_p^+>p^+$ or $\xi_p^-<p^-$. In this case, $\min\{\xi_p^+,\xi_p^-\}=\xi_p^+$,  $\max\{\xi_p^+,\xi_p^-\}=\xi_p^-$, and the ZMP constraint \eqref{zmp} implies that $(1-\frac{\lambda}{\lambda^+})\dot{c}_x\geq 0$ and $(1-\frac{\lambda}{\lambda^-})\dot{c}_x\leq 0$. Thus, $\xi_p^+$ diverges to infinity or $\xi_p^-$ diverges to negative infinity, either of them violates the definition of capturability. As a result, $\xi_p^+\leq p^+$ and $\xi_p^-\geq p^-$ for the case of $\dot{c}_x>0$. Analogously, one can show that $\xi_p^+\geq p^-$ and $\xi_p^-\leq p^+$ for the case of $\dot{c}_x<0$.
\vspace{-10px}
\subsection{Proof of Theorem \ref{thm_bc}}

It suffices to show that the LP problems \eqref{lp1}-\eqref{lp2} is always solvable along the system trajectory starting from any $\x_0\in\mathrm{int}(\Omega)$. At the initial CoM state, \eqref{lp1}-\eqref{lp2} are solvable due to the fact that $\xxi(\x_0)$ is strictly within $\U$.
According to Theorem \ref{lem1}, $|\xi_\lambda-\xi_\lambda^d|$ is strictly decreasing as long as $k_2\geq \epsilon$. Thus, (\ref{ic1}) is satisfied for sufficiently small $\epsilon$. Also, we know from \eqref{ct} that 
$k_2\rightarrow 0$ implies $\eta_p\rightarrow 0$. This indicates that \eqref{ic1'} can be fulfilled provided that $\epsilon$ is sufficiently small. Finally, the proof is completed by noting that $(1-\gamma)(p^--\xi_p)\leq k_1(\xi_p-\xi_p^d)\leq (1-\gamma)(p^+-\xi_p)$ can always be made valid since $\gamma\in (0,1)$.

\end{document}